\documentclass[sigconf,nonacm]{acmart}

\usepackage[utf8]{inputenc}
\usepackage{graphicx,amsmath, multirow}
\usepackage{nicematrix,microtype,booktabs}
\usepackage{makecell} 

\renewcommand{\star}[1]{#1\ensuremath{^*}\kern-\scriptspace}
\newcommand{\tb}[1]{\textbf{#1}}

\begin{document}

\title{End-to-end multi-modal product matching in fashion e-commerce}

\author{Sándor Tóth}
\email{sandor.toth@zalando.de}
\affiliation{
  \institution{Zalando SE}
  \city{Berlin}
  \country{Germany}
}
\orcid{0000-0002-7174-9399}

\author{Stephen Wilson}
\email{stephen.wilson@zalando.ie}
\affiliation{
  \institution{Zalando SE}
  \city{Dublin}
  \country{Ireland}
}

\author{Alexia Tsoukara}
\email{alexia.tsoukara@zalando.ie}
\affiliation{
  \institution{Zalando SE}
  \city{Dublin}
  \country{Ireland}
}

\author{Enric Moreu}
\email{enric.moreu@zalando.ie}
\affiliation{
  \institution{Zalando SE}
  \city{Dublin}
  \country{Ireland}
}

\author{Anton Masalovich}
\email{anton.masalovich@zalando.ie}
\affiliation{
  \institution{Zalando SE}
  \city{Dublin}
  \country{Ireland}
}

\author{Lars Roemheld}
\email{lars.roemheld@zalando.de}
\affiliation{
  \institution{Zalando SE}
  \city{Berlin}
  \country{Germany}
}

\begin{abstract}
Product matching, the task of identifying different representations of the same product for better discoverability, curation, and pricing, is a key capability for online marketplace and e-commerce companies. We present a robust multi-modal product matching system in an industry setting, where large datasets, data distribution shifts and unseen domains pose challenges. We compare different approaches and conclude that a relatively straightforward projection of pretrained image and text encoders, trained through contrastive learning, yields state-of-the-art results, while balancing cost and performance. Our solution outperforms single modality matching systems and large pretrained models, such as CLIP. Furthermore we show how a human-in-the-loop process can be combined with model-based predictions to achieve near perfect precision in a production system.
\end{abstract}


\maketitle

\section{Introduction}

Product matching is a key capability for both marketplace and e-commerce companies, who employ these methods in multiple business areas. On online marketplaces, product matching is used to de-duplicate the assortment and enrich product metadata, thus increasing customer satisfaction when searching for a product or browsing the catalog. Product matching may also support competitive intelligence through automated monitoring of competitors' assortment and prices. Because product matching systems tend to learn strong notions of product similarity, they can furthermore be part of recommendation systems, for example suggesting close substitutes for sold-out products.

We consider the product matching problem on an online marketplace with multiple sellers, each having their own curated assortment. Our task is to merge offers from multiple sellers that correspond to the same product. Although each offer has high quality metadata with product images, title, price, etc, there is a substantial distributional shift among sellers in the visual and text information. For example, product titles may not be sufficiently unique, and product images are created in different styles on each domain: human model poses, camera angles and lighting conditions can differ. This poses a challenge to our high-precision matching system.

A distinctive feature of the fashion market is that visual information is highly important for the customer, and as a result products are often not identified through unique names (beyond brand names). Product matching for fashion items thus has to heavily rely on product both images and text. This makes the fashion product matching problem more similar to person re-identification than matching other product categories \cite{wieczorek20}.

To achieve very high levels of matching precision, human-in-the-loop validation may be added to the machine learning systems. In the matching context, trained humans can validate a subset of the predicted matches and reject invalid predictions. Here we describe how we implemented an optimized human validation process that was able to efficiently reject false positive model predictions, to achieve precision levels required for productive use.

Our paper's main contributions are the following: we propose a simple multi-modal architecture that can cheaply improve the product matching performance over pretrained text models, moreover we compare the two most relevant pretrained visual encoder model families (CLIP and DINO) and surprisingly learn that CLIP based models strongly outperform DINO in image only product matching. Finally, we present an end-to-end production architecture, including human validation steps and show that our method is an efficient solution in industry setting supported by detailed ablation study.

\section{Related work}

While realistic product data is most often multi-modal, containing images and text, the literature on product matching is mostly split on data modality. For textual data, the term entity matching \cite{Christophides2020} is used to describe the task of finding descriptions across datasets that refer to the same real world entity. For images, matching mostly falls under content-based retrieval, where the goal is to find images showing the same object.

Deep learning models achieve state-of-the-art performance in textual product matching and they are often trained as a binary classifier on matching and non-matching concatenated text pairs \cite{Li2020,Yao2022}. Recently, contrastive learning techniques got attention as well, notably for their more efficient use of training data compared to cross encoders trained on the binary classification task \cite{Peeters2022}.

Another relevant area of research is metric learning, where the goal is to learn a mapping of objects to a metric space where the distance between two objects corresponds to how similar they are. Typical use cases are face verification \cite{Schroff2015}, person re-identification \cite{Hermans2017} and fashion retrieval \cite{Goenka2022}. Recent state-of-the-art results are based on fine-tuning pretrained vision encoders to improve metric properties of the embedding space \cite{Ermolov2022}.

Modern computer vision applications rely heavily on large pretrained visual encoders, harnessing the generalization capabilities provided by the large and diverse pretraining dataset. These models work out of the box on any image level (e.g. classification) and pixel level (e.g. segmentation) task. The two most important model class differ in their training objective, these are text-guided pretraining and self-supervised learning. Text guided pretraining uses image captions as a form of supervision to train an image encoder \cite{Mahajan2018}, most models trained in this fashion are variants of CLIP \cite{Radford2021}. It is expected that due to the natural language supervision, general concepts expressed in language are captured by these encoders. An alternative method is self-supervised learning where only images are used as supervision \cite{Chen2020,He2022}, the most prominent model in this group is DINO \cite{Caron2021,Oquab2023} and its variants. Due to the image only training, these models are expected to capture visual features beyond what can be expressed in natural language, thus they are expected to give strong baseline for visual product matching.

\section{Methods}

\subsection{Problem definition}

\begin{table*}[!htb]
    \begin{tabular}{c|c|c|c}
        \tb{Image set}   & \includegraphics[height=40mm]{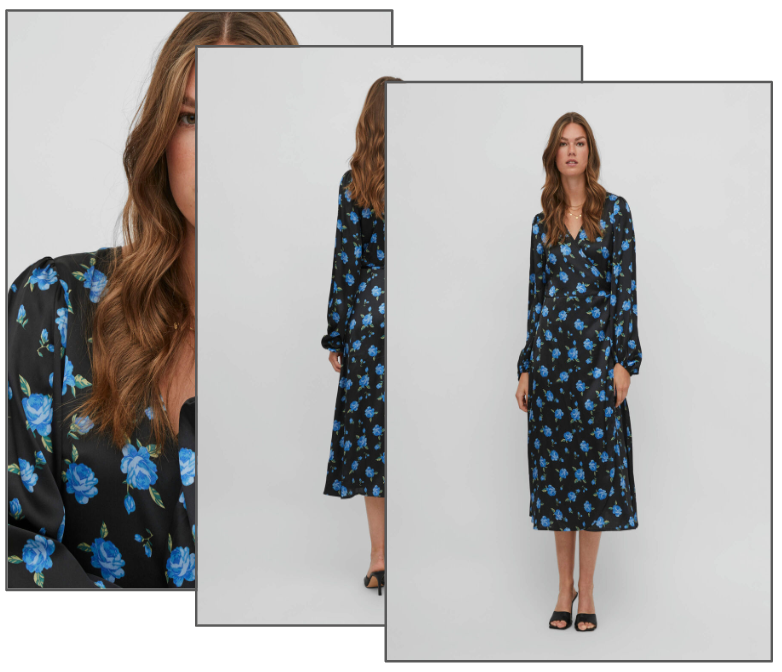} & \includegraphics[height=40mm]{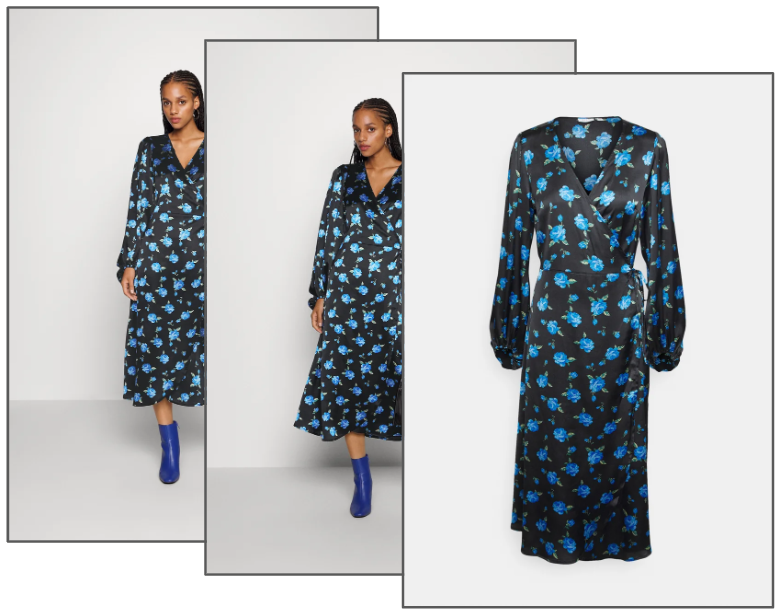} & \includegraphics[height=40mm]{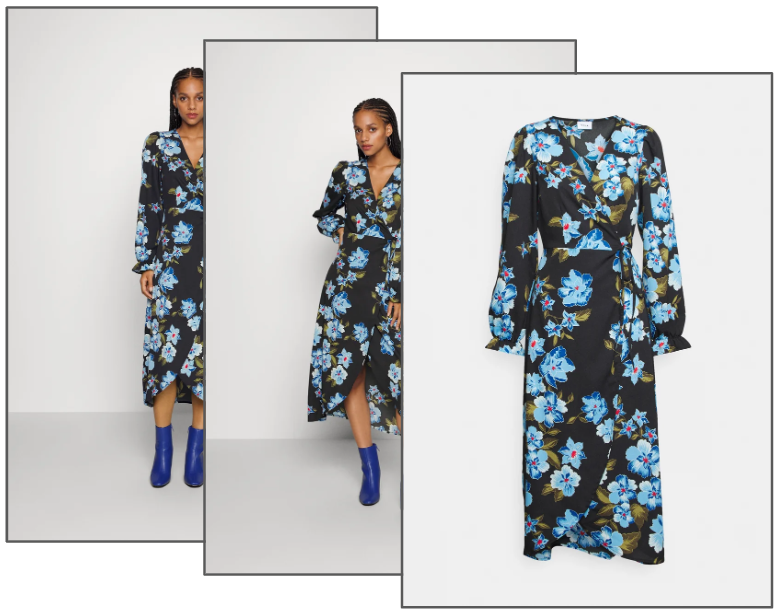} \\ \hline
        \tb{Brand}       & Vila                                       & Vila                                       & Vila                                       \\ \hline
        \tb{Title}       & LONG-SLEEVED Wrap Dress                    & VISASHA MYA - Day dress                    & VITINA WRAP DRESS - Maxi dress             \\ \hline
        \tb{Price}       & €46.9                                     & €44.99                                    & €42.99                                    \\ \hline
        \tb{Num. sizes}  & 6                                          & 7                                          & 7                                          \\ \hline
        \tb{Prod. codes} & 5715362978151, …                           & 5715362978151, …                           & 6582354869017, …                           \\ \hline
    \end{tabular}
    \caption{Sample of offers from our proprietary training dataset: the first two represent the same product from two different domains. The product images are manual screenshots from brand web shops and Zalando's own platform and serves the purpose of illustration.}
    \label{tab:example}
\end{table*}

We consider the problem of matching identical products across multiple domains of representation styles. The different domains can be different sellers within the same fashion marketplace or different online fashion stores on the web. Each domain has several \textit{offers}, where we define an offer as a set of images and textual information representing a given \textit{product}, as in Tab.~\ref{tab:example}. The domain distinction is important, because each domain has unique ways to represent products (e.g. typical model pose on product images, lighting conditions, etc.). The task is to find offers across domains that represent the same product.

\subsection{Dataset and preprocessing}

\begin{table}[!htb]
    \centering
    \begin{tabular}{@{}lcccc@{}}
        \toprule
                            & \tb{train}            & \tb{\begin{tabular}[c]{@{}c@{}}test\\ in-domain\end{tabular}} & \tb{\begin{tabular}[c]{@{}c@{}}test\\ out-domain\end{tabular}} \\\midrule
        \tb{index offer count}    & \multirow{2}{*}{1.5M} & 442k                                                          & 442k                                                           \\
        \tb{query offer count}   &                       & 90k                                                           & 15k                                                            \\
        \tb{positive pair count} & 0.2M                  & 14k                                                           & 6k                                                             \\
        \tb{domain count}   & 4                     & 2                                                             & 2                                                              \\
        \bottomrule
    \end{tabular}
    \caption{Size of the test and train datasets. Index and query correspond to offers belonging to two distinct domains, positive pairs are ground truth matched offer pairs across domains. In-domain and out-domain test datasets share the same index offers.}
    \label{tab:counts}
\end{table}

Our proprietary dataset consists of 2 million offers with mostly studio quality images from a total of 5 domains, as detailed in Tab.~\ref{tab:counts}. Due to copyright reasons, we cannot share this dataset, however similar open source datasets also exist. The largest open source product matching dataset, WDC \cite{Primpeli2019} is comparable in size and contains a broad range of product categories beyond fashion. The largest fashion oriented dataset is DeepFashion2 \cite{Ge2019}, it contains consumer images with more than 0.8M image pairs depicting matching products. The outstanding property of our dataset is its high quality labels, with a manually estimated number of false positives below a few hundred.

To define our product matching problem formally, we assume that there is a set of all offers denoted by $\mathcal{O}$, the set of all products denoted by $\mathcal{R}$ and the ground truth mapping (via labels) from offers to products $C: \mathcal{O} \rightarrow \mathcal{R}$. Denoted by $\mathcal{P} \subset \{ \mathcal{O} \times \mathcal{O} \}$ the set of matching pairs, we then define a matching offer pair $(i,j)$ as:
\begin{equation}
    (i,j)\in\mathcal{P} \Leftrightarrow C(i)=C(j).
\end{equation}
On a high level we can define the matching task as finding a binary classifier that correctly classifies product pairs as match or non-match defined as $f: (i,j)\rightarrow \{0,1\}$. An important property of our dataset is the large number of "lone negatives": offers with no match to any other offer in the dataset, offer $i$ is a lone offer if $C(i)\neq C(j) \forall j\in \mathcal{O}\setminus i$. An interesting question that we want to investigate is whether these lone negatives are useful for training.

Every offer in the dataset has a corresponding set of images with $4.5\pm2$ images on average ($\pm$ denotes standard deviation). During preprocessing we downscale and crop the images to 254x367 (WxH), based on the most common aspect ratio found in our dataset. Some of the image sets contain "packshot" photos (showing the whole product from the best angle without a human model) while the rest of the photos include a human fashion model or specific garment details. The text information contains brand names that are mostly universal across domains with some exceptions:
\begin{itemize}
    \item some brands have multiple sub-brands, e.g. Adidas sub-brands are \texttt{ADIDAS ORIGINALS}, \texttt{ADIDAS TERREX}, etc,
    \item some sub-brand names do not contain the parent brand, e.g. \texttt{Jordan} is part of \texttt{Nike}.
\end{itemize}

As additional text information, we consider only product titles. For fashion items, titles rarely contain uniquely identifying names. Finally, to simplify the introduction of new domains, we only do a minimal preprocessing on the text data. We normalize unicode text in title and brand, then join them into a single text feature.

In addition to image and text data, we also build a vector of numerical features. For these features, we consider the original non-discounted price (typically a standard ``recommended retail price'' directly from the brand) and the number of available sizes per offer as additional features, to build a 3-dimensional numerical feature vector:
\begin{equation}
    v_{num} = \left[n^{size}_i, log(n^{size}_i), log(price_i)\right].
\end{equation}

When creating our test data split, our aim was to measure how our models adapt to new fashion they were not trained on. Both of our test datasets include two domains where the task is to find product matches across these domains. We call the larger domain the index domain and the smaller domain the query domain, similarly how it is used in retrieval literature. We select one domain and use it exclusively for testing. We also pick another 2 domains and split them according to a given date such that offers created before this date are part of the training data and offers created after will be test data. From the resulting 3 test domains we form two test datasets. The in-domain test dataset contains the two domains that are date split from training, thus sharing both domains with training but split in time. The out-domain test dataset contains the query domain that is completely split from training and having the same index domain as the in-domain test set. Testing on the out-domain set enables us to evaluate cross-domain generalization and testing on both test sets enables us to evaluate model generalization over time \cite{Wang2022}. Important to note that both test sets contain a high number of lone negatives: more than 99\% of index offers have no match in the corresponding query set. Detailed count statistics of the datasets are shown in Tab.~\ref{tab:counts}. Our validation dataset is a random split from the remaining training data.

\subsection{High level solution}

We solve product matching in two stages. Since in individual domains all offers are unique with no within-domain matches, we always match offers between two selected domains: index and query. The two steps are:
\begin{enumerate}
  \item encoding: embed all offers with image, text and numerical features into a metric space,
  \item retrieval: iterate over the query offers, search for nearest neighbors within the index set, and select best match candidates whenever sufficiently close neighbors exist.
\end{enumerate}

\subsection{Encoder}

\begin{figure}[!htb]
    \includegraphics[width=\columnwidth]{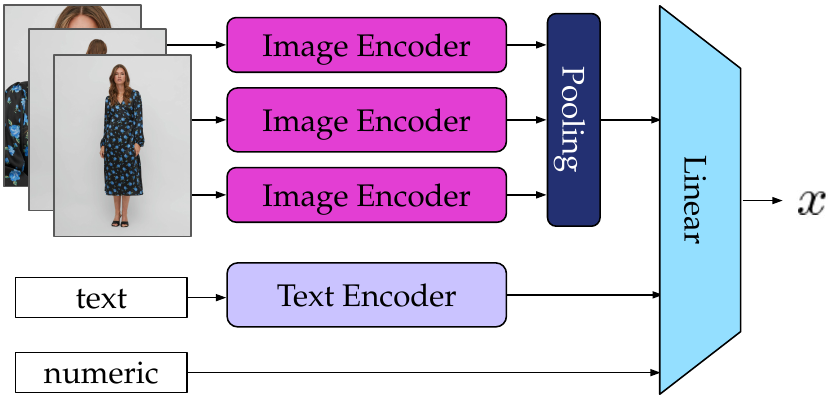}
    \caption{The setup of the multi-modal fashionID encoder.}
    \label{fig:encoder}
\end{figure}

Here we propose fashionID, a multi-modal encoder with pretrained image and text encoder towers, see Fig.~\ref{fig:encoder}. To account for the different number of images per offer, we apply average pooling over the image embeddings of the image set. The resulting image embeddings, text embeddings  and numerical feature vectors are concatenated into a single feature vector. Finally, a linear layer is applied to project this high-dimensional vector into a lower dimensional metric space. This late fusion setup enables us to use any pretrained text and image model and benefit from the low cost of training only the linear projection. Moreover it enables us to adapt newer and larger pretrained image and multi-modal image-text encoders at a small training cost.

\subsubsection{Model training}

To train the model we use the following contrastive loss \cite{Khosla2020} calculated over a mini-batch ($\mathcal{B}$) and using normalized offer embeddings ($v$) of individual offers, some matching and some not:
\begin{equation}
    \mathcal{L} = \sum_{\substack{i\in \mathcal{B} \\ |\mathcal{P}(i)|>0}}-\frac{1}{|\mathcal{P}(i)|}\sum_{j\in \mathcal{P}(i)}\log \frac{\exp(v_i\cdot v_j/\tau)}{\sum_{k\in \mathcal{B}\setminus \{i\}}\exp(v_i\cdot v_k/\tau)},
\end{equation}

where $\tau$ denotes the temperature hyperparameter and $\mathcal{P}(i) \equiv \{j | j\in\mathcal{B} \land (i,j)\in \mathcal{P} \}$ denotes the set of matching offers of offer $i$. We note that this formulation of contrastive loss enables the inclusion of lone negatives to the loss function. Since random sampling would make the number of positive pairs linear in the batch size, and the number of negative pairs quadratic, we use a batch sampling strategy that maximizes the number of positive pairs. To achieve this, during training we uniformly sample product ids and include all offers that correspond to a given product id in the same mini-batch. Also essential for contrastive learning are hard negative samples \cite{Robinson2020}. There are several methods to add hard negatives to the mini-batch, such as offline hard negative mining \cite{Harwood2017}, in-batch memory \cite{Wang2020}, or large mini-batch. Here we made a very important design decision: limiting our training to the linear layer only. This has several advantages. It enable us to precalculate the pooled image and text embeddings, save computation cost and dramatically reduce the GPU memory requirements of our training. This means that we can use large mini-batch sizes without complex and costly multi-GPU training setup. Moreover this removes the need to do explicit hard negative mining for contrastive learning, we can simply train on very large batch sizes. Overall we find this choice a good compromise between the complex requirements of contrastive learning and the usage of large pretrained models.

\subsection{Retrieval}

\begin{figure}[!htb]
    \includegraphics[width=\columnwidth]{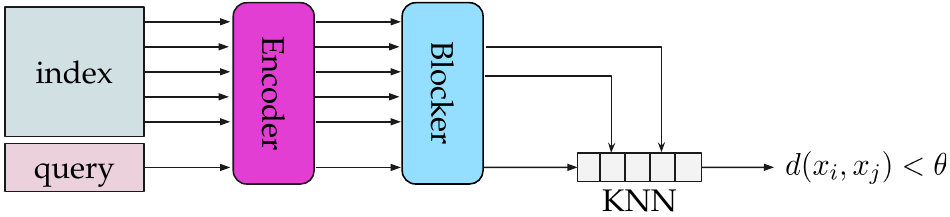}
    \caption{Architecture of the retrieval system.}
    \label{fig:architecture}
\end{figure}

Our retrieval system uses stored fashionID embeddings and searches for nearest neighbors (NN) between query and index table, see Fig.~\ref{fig:architecture}. For this we define the distance between two offers as cosine similarity between the corresponding embeddings. This can be expressed as
\begin{equation}
    d_{i,j} = 1 - v_i\cdot v_j,
\end{equation}
where $v_i$ and $v_j$ are normalized offer embedding vectors. For some applications it is useful to also define an offer similarity score as $s_{i,j}=1-d_{i,j}$ that is larger for more similar products.

Due to the large size of our query and index tables, direct NN search is impractical. A commonly used technique to speed up retrieval is a blocking stage that reduces the search space, the requirement for a good blocking stage is high recall and a high true-negative rate \cite{Steorts2014}. Here we use a fuzzy string similarity metric calculated between query and index brand names and we search for NNs only within brands above a given string similarity threshold. NN search is done using standard algorithms \cite{Johnson2021}. Finally we apply a discriminator to filter out all NN pairs that have distance above a threshold. The threshold for the production system is tuned from calculated precision-recall curves on the test data.

\subsection{Metrics}

We use a set of metrics to evaluate our model both during training and afterwards on the test set. During training we calculate recall at $k$ (R@k) on the validation set; that is, we calculate it as the share of offers where the first $k$ nearest neighbors contain at least one correct match, summing over offers that have at least one ground truth match in the validation dataset. For testing we also measure the precision-recall (PR) curve at $k=1$ by varying the distance threshold (see Fig.~\ref{fig:pr}). For our product matching application both the low recall - high precision region and the high recall low precision region are relevant. The predictions where the expected precision is above a business required threshold can be accepted without human validation. For predictions with lower expected precision, predictions are validated by humans before being accepted. To accommodate both requirements, we use the area under the PR-curve (AUCPR) on the test set as a final metric. This metric accounts for model improvements both at high precision - low recall and at high recall - low precision regions, unlike the equally common $F_1$ score, which ignores model improvements in the high precision region.

\section{Results}

\subsection{Encoder}

To define baseline models, we tested image-only encoders that consisted of a large pretrained image model and average pooling over the image set. As our later analysis confirms, image encoders provide a much stronger baseline than any text encoder. We tested both a CLIP and DINO type pretrained model as baseline:
\begin{itemize}
    \item DINOv2(ViT-G-14), with 1.1B parameters and 1536 output embedding dimension \cite{Oquab2023},
    \item CLIP(ViT-bigG-14) trained on the LAION-2B English subset of LAION-5B, with 1.8B parameters and 1664 output embedding dimension \cite{Radford2021,Ilharco2021,Schuhmann2022,Cherti2023}.
\end{itemize}
We chose the largest pretrained models to assess what is the highest achievable performance with off-the-shelf models.

\begin{table*}[!htb]
    \begin{NiceTabular}{ccccccccccc}
        \toprule
        \multicolumn{2}{c}{\tb{Encoders}}     & \multirow{2}{1.5cm}{\centering \tb{Numerical inputs}} & \multirow{2}{2.0cm}{\centering \tb{Trained linear projection}} & \multirow{2}{*}{\tb{Dim}} & \multicolumn{3}{c}{\tb{in-domain}}                     & \multicolumn{3}{c}{\tb{out-domain}}             \\\cline{1-2} \cline{6-11} 
        \tb{Image}        & \tb{Text}         &                                                       &                              &                           & \tb{AUCPR}       & \tb{R@1}         & \tb{R@3}         & \tb{AUCPR}       & \tb{R@1}  & \tb{R@3}         \\\midrule
        DINOv2(ViT-G-14)  & -                 & -                                                     & -                            & 1536                      & 2.4              & 22.7             & 32.4             & 13.0             & 40.6      & 53.7             \\
        CLIP(ViT-bigG-14  & -                 & -                                                     & -                            & 1664                      & 27.2             & 62.9             & 77.9             & 37.9             & 70.1      & 85.5             \\
        offerDNA          & -                 & -                                                     & -                            & 128                       & 56.9             & 74.0             & 88.5             & \star{\tb{63.9}} & 78.8      & 90.9             \\\midrule
        DINOv2(ViT-G-14)  & -                 & -                                                     & yes                          & 192                       & 38.1             & 63.6             & 83.1             & 53.5             & 73.8      & 88.8             \\
        CLIP(ViT-L-14)    & -                 & -                                                     & yes                          & 192                       & 50.1             & 73.2             & 89.6             & 55.9             & 77.7      & 90.9             \\
        CLIP(ViT-bigG-14) & -                 & -                                                     & yes                          & 192                       & \star{\tb{57.2}} & \star{\tb{77.8}} & \star{\tb{93.0}} & 58.4             & \tb{79.2} & \star{\tb{91.8}} \\\midrule
        -                 & CLIP(ViT-bigG-14) & -                                                     & yes                          & 192                       & 0.6              & 11.1             & 23.5             & 2.6              & 17.9      & 36.0             \\
        offerDNA          & CLIP(ViT-bigG-14) & yes                                                   & yes                          & 192                       & 63.8             & 79.6             & 93.0             & \tb{64.3}        & 80.6      & 92.5             \\
        CLIP(ViT-bigG-14) & MUSE              & yes                                                   & yes                          & 192                       & 63.9             & 83.4             & 95.0             & 62.3             & 81.4      & 92.3             \\
        CLIP(ViT-bigG-14) & CLIP(ViT-bigG-14) & -                                                     & yes                          & 192                       & 64.9             & 83.7             & 94.9             & 62.6             & 81.9      & 92.5             \\
        CLIP(ViT-bigG-14) & CLIP(ViT-bigG-14) & yes                                                   & yes                          & 192                       & \tb{66.1}        & \tb{84.2}        & \tb{95.2}        & 63.3             & \tb{82.1} & \tb{92.6}        \\
        \bottomrule
    \end{NiceTabular}
    \caption{Comparison of test performance for raw embeddings of pretrained and fine tuned models and when combined with a trained linear projection. Best overall values are denoted by bold. \star{} best image-only performance}
    \label{tab:results}
\end{table*}

To illustrate the relative performance of the methods described here, we also include results for a proprietary image encoder with a stack of ConvNext Tiny  with 28M parameters \cite{Liu2022}, set attention \cite{Lee2019} and average pooling that was specifically trained to encode offer image sets to 128 dimensional metric embeddings, the model we will call offerDNA. The offerDNA model, trained with offline hard negative mining \cite{Harwood2017}, was our previous state-of-the-art and serves as a demonstration of what results can be achieved with a small end-to-end trained model. The three baseline results are shown in Tab.~\ref{tab:results}. The much smaller offerDNA model outperforms off-the-shelf pretrained models by a large margin in the AUCPR metric. Interestingly, the difference in recall is smaller, which implies that increased AUCPR is due to the increased precision of the fine-tuned model at low recall. Surprisingly, we also observe that CLIP outperforms DINO by a large margin while having similar parameter count.

\begin{figure}[!htb]
    \includegraphics[width=\columnwidth]{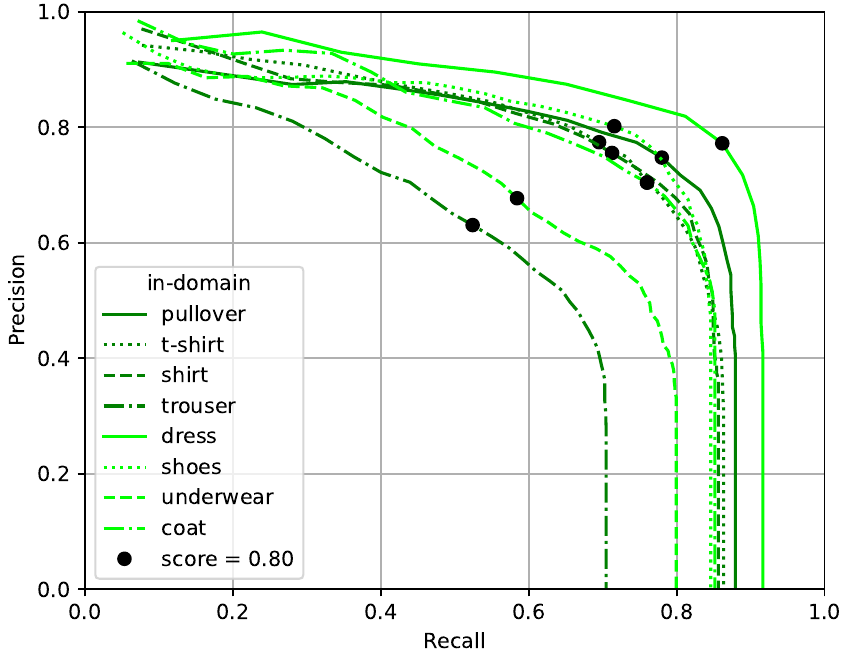}
    \caption{Precision-recall curves for fashionID evaluated on the most common fashion categories of the in-domain test set. Black dots denote the precision - recall values at the fixed similarity score of 0.80 per category.}
    \label{fig:pr}
\end{figure}

As a best model, we trained the fashionID encoder using CLIP(ViT-bigG-14) as image and text encoder with frozen weights. We trained a linear projection with output dimension of 192 and 0.6M parameters (input dimension of the projection is $2\cdot 1664$, the concatenated length of text and image embeddings). For training and further ablation studies we used an AdamW optimizer with a relatively large learning rate of $10^{-3}$ and a temperature of $\tau=0.06$ in the contrastive loss term, close to the published value of 0.1 in a similar contrastive learning study \cite{Khosla2020}. We also filtered out lone negatives from the training data, detailed analysis on this follows below. Typical training time was 10 minutes for 50 epochs on a single NVIDIA A100 GPU. Throughout our experiments we observed no overfitting on the validation set due to the low number of fitted model parameters. The model outperformed all baselines, including the custom fine-tuned offerDNA model and achieved top score in almost all metrics. FashionID precision - recall curves calculated on the in-domain test data show fairly uniform performance on the top fashion categories, see Fig.~\ref{fig:pr}, with the only exception being trousers. We confirmed by manual inspection that trousers have a low level of visual variability and there are a high number of identical looking trouser offers where different products are differentiated only by small variation in shape and cut.

To better understand how each data modality contributes to the performance of fashionID, we retrained the model using image- and text-only modality. The results confirm our expectation that visual information is the most important modality for fashion product matching, where the trained text-only fashionID model is outperformed even by  raw CLIP embeddings of images. We also did an additional ablation, removing the numerical features and training an image and text only fashionID. We found that the 3 numerical features give around to 1 percentage point AUCPR improvement on both in-domain and out-domain test sets. We also tested a pretrained multilingual universal sentence encoder (MUSE) \cite{Yang2019, Reimers2019} instead of the CLIP text encoder. Even though the multilingual encoder is expected to better encode some of our German language product titles, we did not observe improvement over CLIP.

\begin{figure}
    \includegraphics[width=\columnwidth]{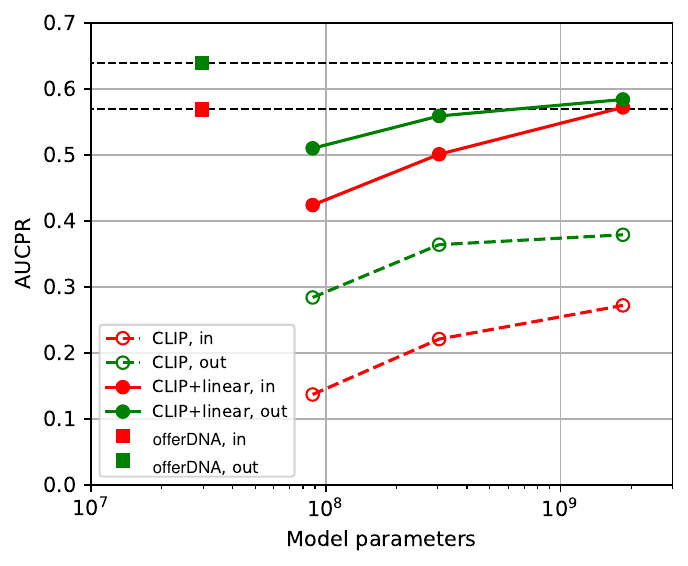}
    \caption{Comparison of matching performance as a function of pretrained model parameter count. We show the parameter count of (frozen) CLIP encoders with and without linear projection and the small fully fine-tuned offerDNA, "in" / "out" denotes results on the in-domain and out-domain test sets, respectively.}
    \label{fig:params}
\end{figure}

We conducted further studies using only the visual modality to better understand the different performance trade-offs. We compared image encoders with different pretraining objective and found that CLIP consistently outperforms DINOv2 both as raw embeddings and when combined with a trained linear projection. When comparing pretrained CLIP with increasing parameter count (ViT-B-32, ViT-L-14 and ViT-bigG-14), we found substantial performance increase even when we keep the final embedding dimension constant, see Fig.~\ref{fig:params}. This strong performance increase as a function of model capacity holds for both the raw embeddings and when combined with trained linear projection. This is expected, as larger CLIP models are able to capture more of the relevant features on the image that we then project out to the 192 dimensional output via the linear projection. Moreover, we found that although the small offerDNA model outperforms the frozen models, the difference in performance is not substantial, test results on the in-domain dataset are nearly identical, showing the potential of using off-the-shelf models for less demanding computer vision applications.

\begin{figure}[!htb]
    \includegraphics[width=\columnwidth]{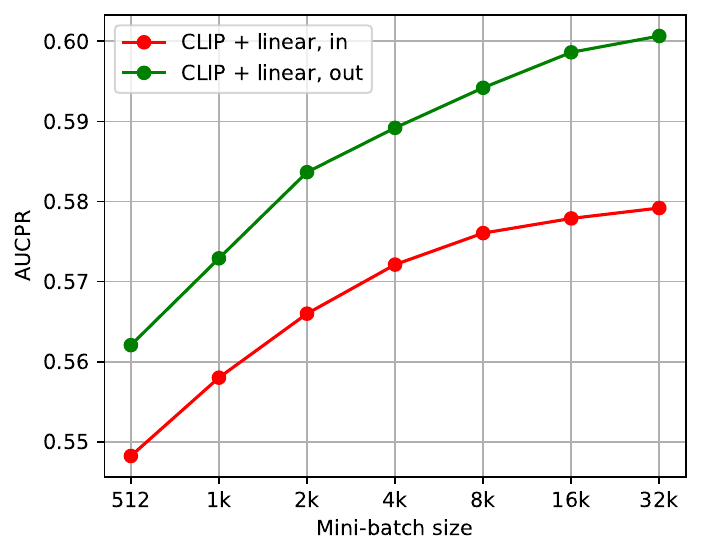}
    \caption{CLIP(ViT-bigG-14) image encoder with linear projection performance as a function of training mini-batch size.}
    \label{fig:batchsize}
\end{figure}

To elucidate the importance of batch size, we trained image-only fashionID models (frozen image encoder + trained projection) with different batch sizes up to 32k. We found strong AUCPR improvement with increasing batch size, see Fig.~\ref{fig:batchsize}. This confirms our hypothesis, that large batch size is essential for contrastive learning. Interestingly when we further increased the batch size, the training became less stable with large fluctuations of the loss during training even when we decreased the learning rate. Thus for all consecutive training we fix the batch size to 16k.

\begin{figure}[!htb]
    \includegraphics[width=\columnwidth]{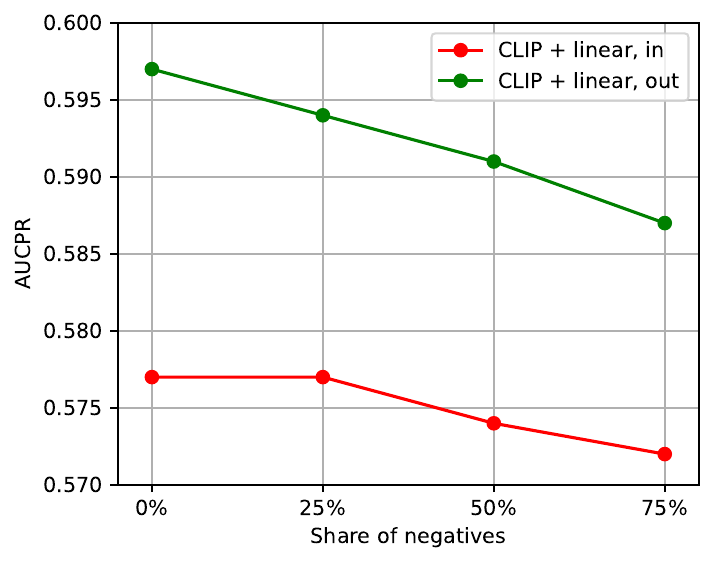}
    \caption{Matching performance as a share of negatives used in training the linear projection on frozen CLIP(ViT-bigG-14) image encoder for in-domain and out-domain test sets.}
    \label{fig:negtives}
\end{figure}

Finally, we studied the effect of lone negatives on model training. Since there is a high number of lone negatives in our test set, our original hypothesis was that our training data should similarly contain lone negatives for best extrapolation to the test set. To find out the role lone negatives playing in training, we gradually downsampled the number of lone negatives in the training set and retrained the image-only fashionID. Surprisingly, we found that lone negatives actually have a negative effect on the results, see Fig.~\ref{fig:negtives}. The performance increase we get by removing lone negatives is approximately equal to the performance increase from the larger effective batch size when counting only positive pairs in the batch. This suggests that the lone negative examples have no contribution in our contrastive learning setup; we hypothesize that this is due to lone negatives not exhibiting fundamentally different characteristics than matched products. We also tested different output embedding dimensions and found that model performance increases sharply until 192 with slow decrease in AUCPR beyond.

\begin{table}[!htb]
    \centering
    \begin{tabular}{@{}lccc@{}}
        \toprule
                                              & \tb{Num. layers} & \tb{0}    & \tb{1}    \\ \midrule
        \multirow{2}{*}{\tb{validation set}}  & R@1             & 88.8      & \tb{90.1} \\
                                              & R@3             & 95.8      & \tb{96.4} \\ \midrule
        \multirow{2}{*}{\tb{test in-domain}}  & R@1             & \tb{77.8} & 77.6      \\
                                              & R@3             & 93.0      & \tb{93.4} \\ \midrule
        \multirow{2}{*}{\tb{test out-domain}} & R@1             & \tb{79.2} & 78.9      \\
                                              & R@3             & \tb{91.7} & 91.6      \\
        \bottomrule
    \end{tabular}
    \caption{Comparison of different number of hidden layers in the trained shallow MLP using CLIP(ViT-bigG-14) image embeddings as input, 0 hidden layer corresponds to a linear projection, 1 hidden layer is the simplest multi-layer perceptron.}
    \label{tab:layers}
\end{table}

To better understand the generalization capability of fashionID, we replaced the linear projection with a multilayer perceptron (MLP). For a one hidden layer setup, we used a 256 dimensional hidden layer with ReLU activation function. We found, as expected, that the increased capacity of MLP over the linear projection improved the validation set performance without overfitting it, see Tab.~\ref{tab:layers}. At the same time the linear projection slightly outperformed the MLP both on the in-domain and out-domain test set, showing that the linear projection maintains the original broad generalization capabilities of the pretrained CLIP encoder.

\section{Human-in-the-loop}

\begin{figure}[!htb]
    \includegraphics[width=\columnwidth]{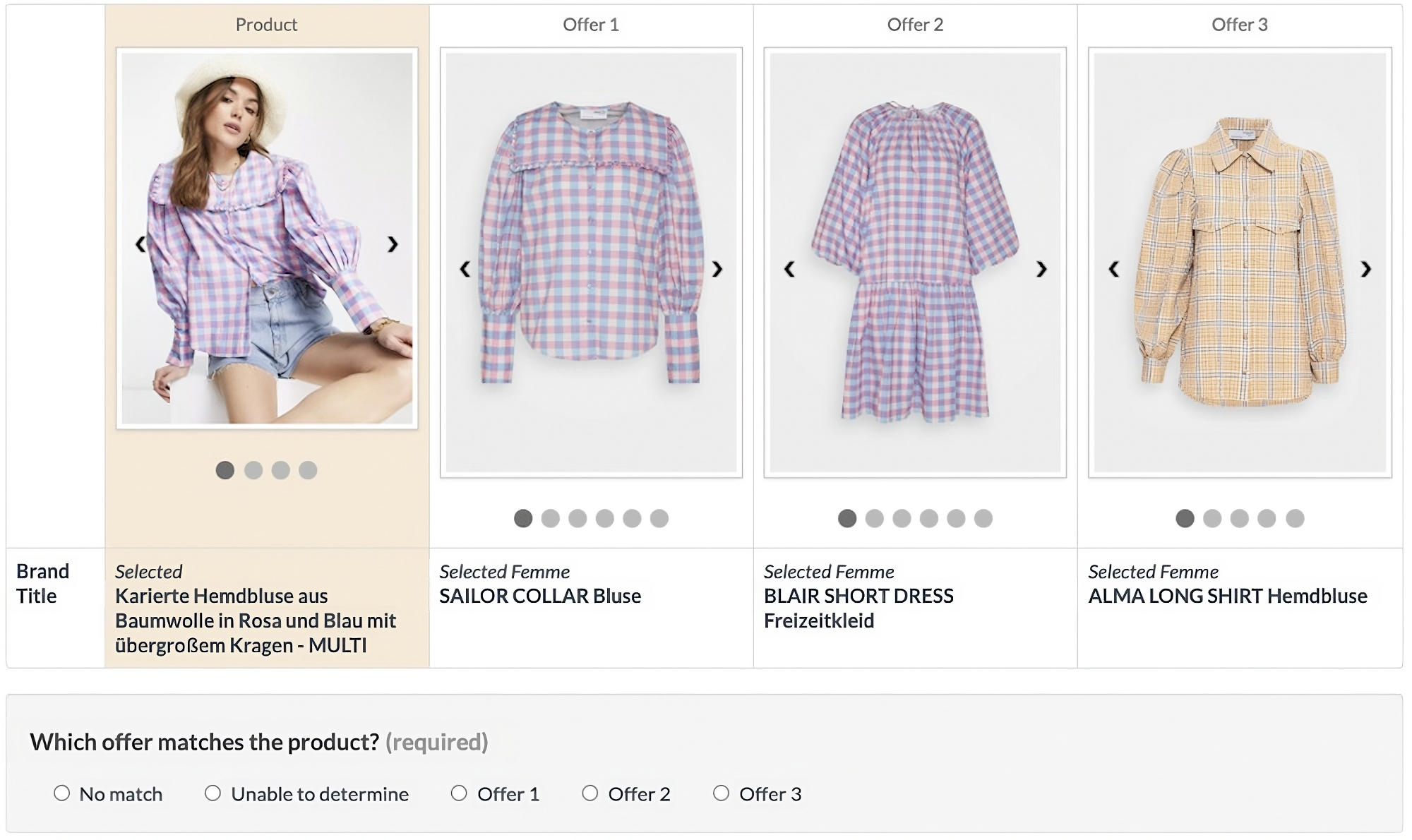}
    \caption{The final UI, created for human validation showing a product on the left and the three ML predicted nearest neighbors.}
    \label{fig:ui}
\end{figure}

If the matching application requirements cannot be met by the best machine learning model, it is possible to further improve performance using human-in-the-loop (HITL). In general, human workers can increase both recall and precision. To increase precision in the product matching use case, trained humans can validate a subset of the predicted matches and reject invalid predictions. To increase recall, humans would need to execute themselves an equivalent of a product search, looking for new matches from the high number of available offers. Due to the high number of negative pairs and the additional requirement of a search capability, increasing recall via HITL is impractical. In the following we describe how we implemented an optimized human validation step that increases model precision. At high level, human validators are shown a visual and text comparison of multiple predicted matches and their task is to vote if one of the presented offers match the given product. We conducted multiple experiments to optimize many aspects of the task using fast iterations where each time 100 rows were manually validated by this paper's authors. Afterwards we used human experts to collect validated data on a larger dataset and eventually HITL became part of the production matching system.

For the human validation experiments we sampled from a fixed set of model-predicted matches including up to 3 nearest neighbors. We measured final precision and recall for each validation setup and chose the setup that produced the highest final recall at the required precision level of 90\%. After several iterations, we arrived at the interface shown in Fig.~\ref{fig:ui}. Our main learnings are:

\begin{enumerate}
    \item it is more efficient to hide some information that is difficult to interpret for validators. Adding information such as color, price or the list of sizes did not improve validation performance, despite being helpful as input features for the ML model,
    \item showing top three nearest neighbors and allowing the validator to choose the correct one increased recall when compared to showing only the first nearest neighbor,
    \item if high precision is required, providing validators with multiple product photos that include close up images is beneficial,
    \item giving feedback to validators brought big performance improvements, we conducted several training rounds each time reviewing the common errors.
\end{enumerate}

\begin{table}[ht]
    \centering
    \begin{tabular}{@{}lcc@{}}
        \toprule
        Experiment                   & v1       & v2       \\ \midrule
        Input model precision         & 0.162    & 0.285    \\
        judgments per row            & \multicolumn{2}{c}{3} \\
        aggregation rule             & \multicolumn{2}{c}{majority vote} \\
        Number of rows               & 14453    & 3151     \\ \midrule
        TPR                          & 0.794(5) & 0.831(8) \\
        FNR                          & 0.206(5) & 0.169(8) \\
        TNR                          & 0.982(1) & 0.978(2) \\
        FPR                          & 0.018(1) & 0.022(2) \\ \midrule
        $LR+$                        & 44(2)    & 38(3)    \\
        HITL output precision        & 0.896(4) & 0.937(5) \\
        predicted HITL precision     & -        & 0.946(7) \\
        \bottomrule
    \end{tabular}
    \caption{Human validator confusion matrices and relevant coefficients measured on ML matched prediction produced by two different ML models, for details see text. Numbers in brackets denote standard deviation on the last digit evaluated via bootstrapping.}
    \label{tab:hitl}
\end{table}

Once we finalized the UI, we compared the performance of dedicated (expert) workers to untrained crowdsourced validators and found that even after applying common consensus-voting rules with 5 votes per row, we couldn't reach the required precision with crowdsourced workers. We also found that at large enough scale dedicated workers are even more cost effective than crowdsourced workers and provide a stable and reliable solution, something that is required for using HITL in production settings. With our dedicated workers, we collected a bigger validated dataset with 3 validator votes per row. We used this data and simulated different judgment aggregation rules and evaluated against the known ground truth product ids. The collected data corresponds to the output of two different matching models validated under the same conditions, for results see Tab.~\ref{tab:hitl}. The v2 prediction output dataset with average model precision of 29\% (the precision is relatively low, because we calculate average precision over pairs up to 3rd neighbor) corresponds to our custom fine-tuned offerDNA model, while the v1 dataset with 16\% average precision corresponds to an earlier model iteration.

Interestingly, we observed that the output precision of HITL may vary with the output precision of the underlying ML matching model. To obtain a heuristic for the overall precision from our system including HITL, we have found the full confusion matrix data from human validation to be informative. To demonstrate this, we calculated the whole HITL confusion matrix from experiment v1, see in Tab.~\ref{tab:hitl}. Here we took aggregated judgments using a simple majority rule: a row is a valid match if it gets more than one positive vote out of the three collected judgments, finally we compared the aggregated judgment to ground truth labels. Now, if we use the known average ML model precision, we can predict the output precision of the human validation using the following formula:

$$P_{hitl} = \left\{1 + \frac{1/P_{model} - 1 }{LR+}\right\}^{-1},$$

where $LR+=TPR/FPR$ is the positive likelihood ratio for the validation process, $P_{hitl}$ and $P_{model}$ are the final and the ML output data precision, respectively. Using this formula and the observed $LR+$ value calculated from v1 validation run, we predict validation output precision for the v2 run (using only the known model precision from the v2 experiment) and we get $\hat{P}_{hitl}=0.946(7)$ that agrees with the observed output precision within 2$\sigma$. Thus we showed that input data precision impacts the human validation performance and in our case the human validation process is best characterized by TPR and FPR values instead of an expected fixed output precision.

\section{Conclusions}

In this paper we showed that a state-of-the-art, large-scale, production-grade multi-modal product matching system can be built based on large pretrained image and text models. We demonstrated that using late fusion via a linear projection layer trained with large batch contrastive learning can achieve both high retrieval performance in a narrow domain, such as fashion, and at the same time can generalize to shifts in the data distribution, as we demonstrated on our out-domain test data. The advantage of this method is fast and cheap experimentation where a full training takes less than 10 minutes on a single GPU once input text and image embeddings are precalculated. As a more general takeaway, we found that CLIP encoders outperform both DINOv2 on visual modality and multilingual universal sentence encoders on text modality on our product matching dataset. Importantly, most of the product matching performance comes from matching semantic information related to fashion on the product images, while the text information is secondary. This is in strong contrast to product matching in other categories, such as electronics, books, etc., where textual features are more informative.

Overall our example suggests that computer vision applications in industry are getting more heavily reliant on large pretrained models even without fine-tuning. As computation cost is going down, deploying these models directly in production enables savings in training cost and complexity, while the pre-calculated raw image embeddings can be stored and reused as a cost efficient drop-in replacement for images in many ML applications.

\bibliographystyle{ACM-Reference-Format}
\bibliography{library-filtered}

\end{document}